\documentclass[journal]{IEEEtran}

\usepackage{amsmath}
\usepackage{amssymb}
\usepackage{amsmath}
\usepackage{bbding}

\usepackage{multirow}
\usepackage{setspace}
\usepackage{color,subfigure}
\usepackage{mathrsfs}
\usepackage{graphics}
\usepackage{epsfig}
\usepackage{url}
\usepackage{booktabs}       
\usepackage{float}

\begin{document}
 
\title{Multi-stage Object Detection \\ with Group Recursive Learning}
 
\author{Jianan~Li,
        Xiaodan~Liang,
        Jianshu~Li,
        Tingfa~Xu,
        Jiashi~Feng,
        and Shuicheng~Yan,~\IEEEmembership{Senior~Member,~IEEE}
\thanks{Jianan~Li and Tingfa~Xu are with School of Optical Engineering, Beijing Institute of Technology University, China. 
Xiaodan~Liang is from Sun Yat-Sen University, China.
Jianshu~Li, Jiashi~Feng and Shuicheng~Yan are from Department of Electrical and Computer Engineering, National University of Singapore.
}
}


\maketitle

\begin{abstract}
Most of existing detection pipelines treat object proposals independently and predict bounding box locations and classification scores over them separately.
However, the important semantic and spatial layout correlations among proposals are often ignored, which are actually useful for more accurate object detection. In this work, we propose a new  EM-like group recursive learning approach to iteratively refine object proposals by incorporating such context of surrounding proposals and provide an optimal spatial configuration of  object detections. In addition, we propose to incorporate the weakly-supervised object segmentation cues and region-based object detection into a  multi-stage architecture in order to fully exploit the learned segmentation features for better object detection in an end-to-end way. The proposed architecture consists of three cascaded networks which respectively learn to perform weakly-supervised object segmentation, object proposal generation and recursive detection refinement. Combining the group recursive learning and the multi-stage architecture provides competitive mAPs of $78.6\%$ and $74.9\%$ on the PASCAL VOC2007 and VOC2012 datasets respectively, which outperforms many well-established baselines~\cite{girshick2015fast}~\cite{ren2015faster} significantly.
\end{abstract}

\section{Introduction}
Object detection is a fundamental problem in computer vision research. In  recent years, remarkable progress has been made in object detection~\cite{NIPS2012_4672,NIPS2012_4562,NIPS2012_4770,NIPS2014_5513}, arguably benefited from the rapid development of deep neural network based methods~\cite{NIPS2014_5418,NIPS2013_5207,girshick2013rich,girshick2015fast,redmon2015you,NIPS2015_5644}. Among them, one of the most influential methods is the R-CNN framework~\cite{girshick2013rich} which performs CNN-based classification on the object proposals produced by various methods (\emph{e.g.}~\cite{uijlings2013selective}~\cite{zitnick2014edge}). As two examples of improvement upon R-CNN, Fast R-CNN~\cite{girshick2015fast} learns a convolutional feature map from the entire  image  before extracting features to classify each proposal, and faster R-CNN~\cite{ren2015faster} combines Region Proposal Network (RPN) and Fast R-CNN with shared convolutional layers. Those two variants  both bring compelling accuracy and efficiency enhancement for object detection. However, existing R-CNN based methods make predictions for each proposal independently, although surrounding proposals of the same object can provide useful information to refine the proposal location to better cover the object.  Moreover, they do not consider segmentation cues which are beneficial for better localizing the objects. In this paper, we aim to further enhance object detection by adopting two strategies, \textit{i.e.} multi-stage network cascades and group recursive learning for detection refinement.

\begin{figure*}
	\centering
	\includegraphics[scale=0.45]{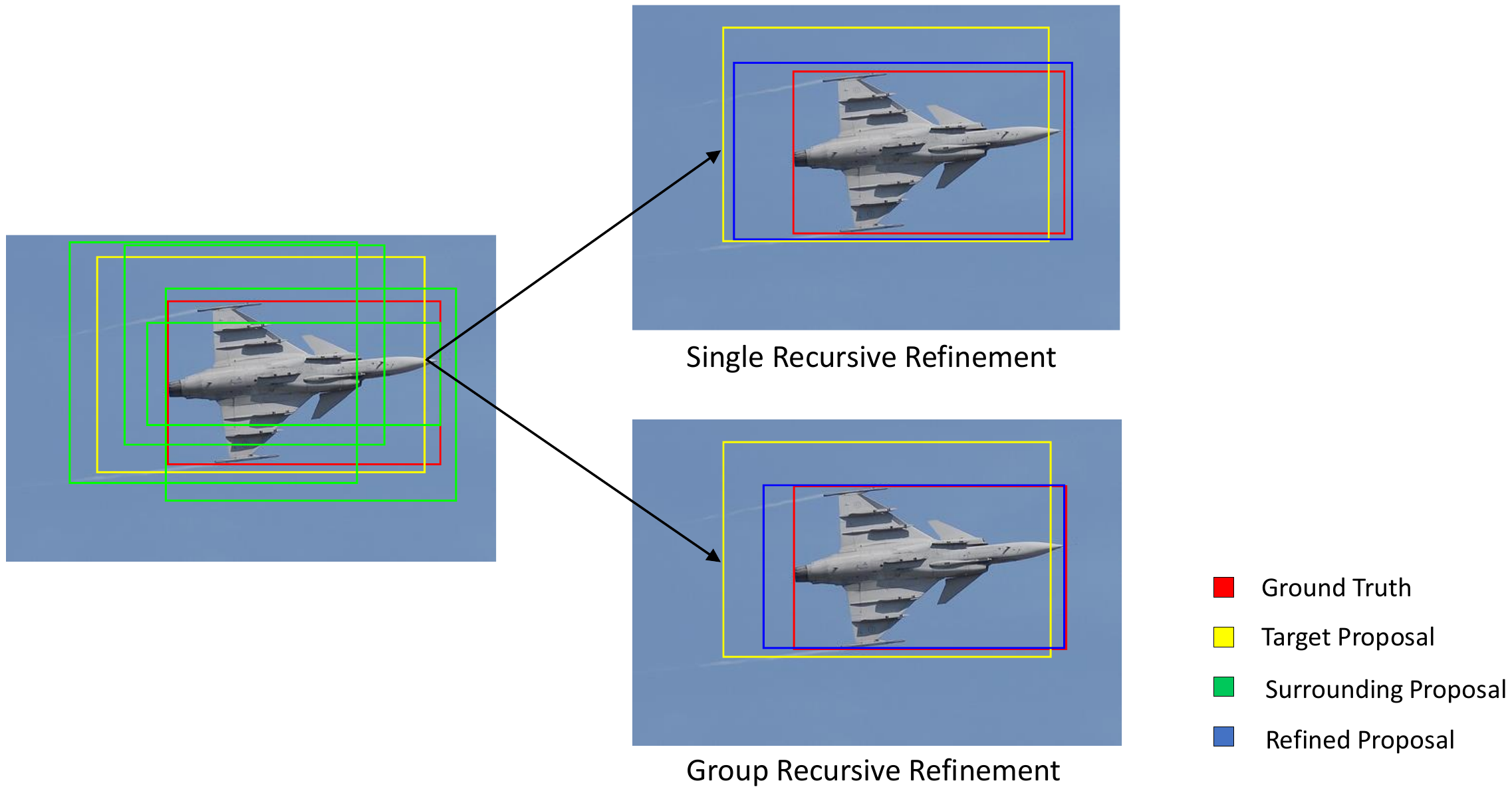}
	\caption{Illustration of group recursive refinement. The red bounding box represents the ground truth location of an airplane. The yellow rectangle and the green rectangles denote the target object proposal to be refined and its surrounding object proposals of the same object. The blue rectangles represent the refined bounding box locations of the target object proposal. Compared to the refined location produced by regressing the target object proposal singly, a more accurate bounding box which tightly encloses the ground truth can be obtained from the group recursive refinement thanks to the useful location cues provided by multiple proposals.}
	\label{fig:proposals}
	\vspace{-3mm}
\end{figure*}

\paragraph{Multi-Stage Network Cascades} Object detection aims to tightly localize objects of particular categories in an image, while semantic segmentation  aims to predict the category label for every pixel of the image. Although the two tasks are typically addressed separately, we argue that the features learned for  semantic segmentation tasks could provide valuable cues for more accurately localizing objects~\textemdash~especially for the ones with small scale or occlusion. Therefore, we propose a multi-stage network cascades architecture to jointly perform weakly supervised semantic segmentation and object detection. The proposed architecture consists of three cascaded networks.
The first network is for weakly supervised segmentation and learns  specific semantic segmentation features from the entire image. The second network generates  object proposals by considering both the convolutional features and the produced  segmentation features. Better object proposals can thus be generated as foreground and background can be better distinguished using the segmentation related cues. Since there exit large variations in the initial locations of the produced proposals, it is usually hard to make precise predictions for some of the proposals independently with only one step. Thus the third network refines detections recursively based on  object proposals produced in the previous stage and global dependency among multiple proposals. 
In this cascade way, the underlying segments from the semantic segmentation task which can provide local cues for better localization can be inherently integrated for object proposal generation and bounding box prediction. Moreover, precise predictions can be progressively obtained through recursive refinement using the global cues from multiple proposals.

\paragraph{Group Recursive Learning} Most existing approaches for object detection perform category prediction for each object proposal independently without considering the proposals in the vicinity. However, the mutual information among a group of neighboring proposals is quite valuable for getting more accurate detection results. As illustrated in Figure~\ref{fig:proposals}, although all of the object proposals have a large overlap with the ground truth, their relative locations to the ground truth and the semantic regions covered are significantly different. Some of the proposals are distant from the ground truth. It becomes difficult for the network to make precise predictions independently with such rough locations. One can observe that for a specific proposal, its surrounding proposals cover different parts of the object. They can provide useful cues to refine the proposal for better concentrating around the actual objects of interest. \\

Following the above intuition, we propose a group recursive learning approach to progressively refine object detection results in an expectation-maximization like manner. More concretely, in the E-step, given initial detection results, our proposed approach further refines each proposal by taking into consideration the surrounding proposals which have large overlap with the proposal of interest. These proposals are considered so that a \emph{group} is formed. All the proposals within the same group collectively refine the proposal of interest to more precise locations. In the M-step, the likelihood of proposals being close to the corresponding ground truth bounding boxes is maximized through the learning process which provides more precise location predictions. This proposed recursive learning procedure is performed iteratively until optimal predictions are achieved.

\paragraph{Contributions} To summarize, we make the following contributions. (1) We develop a unified multi-stage network cascades architecture that is capable of leveraging semantic segmentation features for object detection. (2) We introduce an EM-like group recursive learning approach to iteratively refine detection results and minimize the offsets between object proposals and the ground truth step by step considering the global dependency among multiple proposals. (3) Our detection architecture achieves competitive mAPs of $78.6\%$ and $74.9\%$ on VOC2007 and VOC2012 detection challenges~\cite{everingham2010the} respectively, which outperforms many well established baselines significantly.


\section{Related Work}
In recent years, several works have proposed to incorporate segmentation techniques to assist object detection in different ways. For example, Parkhi \emph{et al.}~\cite{parkhi2011truth} improved the predicted bounding box with color models from predicted rectangles on cat and dog faces. Dai \emph{et al.}~\cite{dai2012learning} proposed to use segments extracted for each object detection hypothesis to accurately localize detected objects. Other research has exploited segmentation to generate object detection hypothesis for better localization. Segmentation was adopted as a selective search strategy to generate the best locations for object recognition in~\cite{van2011segmentation}. Arbelaez \emph{et al.}~\cite{arbelaez2014multiscale} proposed a hierarchical segmenter that leverages multiscale information and a grouping algorithm to produce accurate object candidates. Instead of using segmentation for better localizing detections, Fidler \emph{et al.}~\cite{fidler2013bottom} took advantage of semantic segmentation results~\cite{carreira2012semantic} to more accurately score detections. In this work, we propose a unified framework to incorporate semantic segmentation features for both object proposal generation and better scoring and localizing detections. In addition, a group recursive learning strategy is employed to recursively refine the scores and locations of the detections, thus achieving more precise predictions.

\section{Overview on Multi-stage Object Detection Architecture}
\begin{figure*}
	\centering
	\includegraphics[scale=0.5]{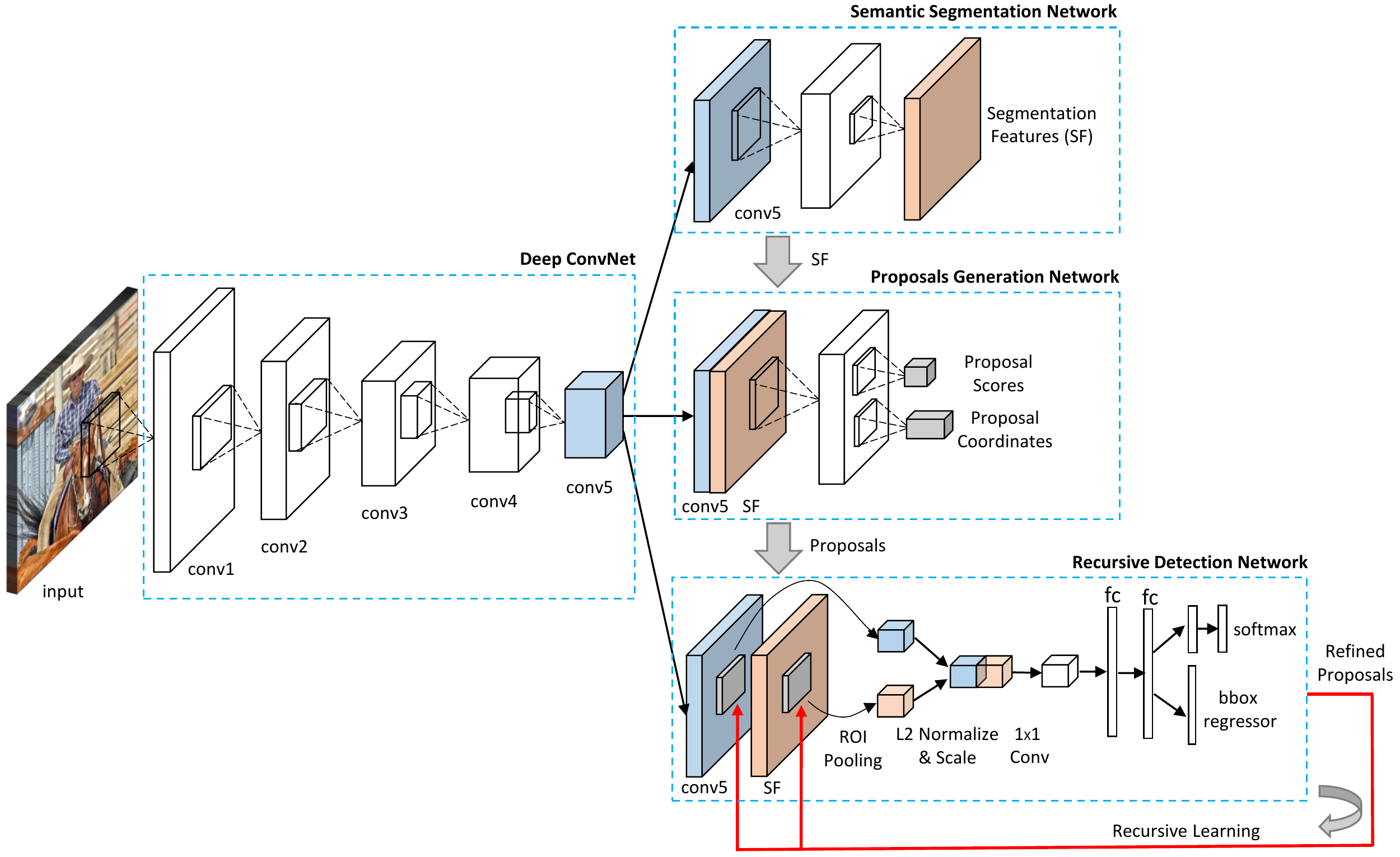}
	\caption{Detailed architecture of the proposed framework. The whole input image is first fed into several convolutional layers and max pooling layers to generate the shared convolutional feature maps. The semantic segmentation network takes as input the shared feature maps and further computes the semantic segmentation feature maps for the input image through several convolutional layers. These computed feature maps are concatenated with the shared convolutional feature maps, forming the input of the proposal generation network to generate object proposals. For each produced proposal, the recursive detection network extracts a descriptor with fixed resolution from both types of feature maps using ROI pooling~\cite{girshick2015fast}. Each descriptor is L2-normalized, concatenated, scaled, and dimension-reduced ($1\times1$ convolution) to produce a fixed-length feature descriptor of size $512\times7\times7$, which is fed into two fully connected layers to predict the confidences of all categories and the bounding box offsets. In addition, a group recursive learning scheme is performed to refine the bounding box locations and classification scores with multiple iterations. In each iteration, the bounding box locations for each proposal are refined by the predicted bounding box offsets and are further updated using the locations of its surrounding proposals of the same object through group refinement for more precise locations.}
	\label{fig:framework}
	\vspace{-4mm}
\end{figure*}

Our proposed object detection architecture consists of a cascade of multiple CNN networks, each  of which focuses on a specific task, \emph{i.e.}, weakly-supervised semantic segmentation, proposal generation and recursive detection refinement respectively. The three networks share convolutional features learned from the entire image. 
Details about the proposed architecture are  shown in Figure~\ref{fig:framework}. The input image first passes through several convolutional  and max pooling layers to produce convolutional feature maps. Then the semantic segmentation network learns semantic segmentation features for the entire image from the convolutional feature maps. The produced features are then fed into the proposal generation network to generate candidate object proposals. Finally, the recursive detection network  iteratively refines the scores and locations of generated object proposals 
via a group recursive learning strategy. In the following subsections, we explain the multi-stage network cascades, group recursive learning scheme and testing phase with more details.

\subsection{Multi-Stage Network Cascades}
Object detection and semantic segmentation are two closely related tasks. The segments extracted for each object proposal can provide useful local cues (\emph{e.g.}, object boundaries) for better object localization. In order to incorporate semantic segmentation cues to assist object detection, we introduce the multi-stage network cascades architecture to jointly perform  weakly-supervised semantic segmentation and object detection, in order to learn better image representations for object detection.  
\subsubsection{Weakly-Supervised Semantic Segmentation Network}
For the semantic segmentation network, we use the semantic segmentation-aware CNN model adopted in~\cite{gidaris2015object} which is trained for the class-specific foreground segmentation task based on a Fully Convolutional Network~\cite{long2015fully}. To avoid using additional segmentation annotations, the network is trained to predict class specific foreground probabilities in a weakly supervised manner with only the provided bounding box annotations for the detection task. The artificial foreground class specific segmentation masks are created using bounding boxes annotations. Specifically, the ground truth bounding boxes of an image are projected on the last hidden layer of the Fully Convolutional Network. The ``pixels'' inside the projected boxes are labeled as foreground while the rest are labeled as background. This process is performed independently for each class. After the Fully Convolutional Network has been trained on the class-specific foreground segmentation task, we drop the last classification layer and extract the convolutional feature maps output by the last convolutional layer as semantic segmentation features for the input images.
\subsubsection{Proposal Generation Network}
Based on the computed feature maps of the input image, the proposal generation network aims to produce a set of object proposals, each of which has a predicted objectness score. Following the Region Proposal Network (RPN) proposed in~\cite{ren2015faster}, the proposal generation network is structured with a convolutional layer followed by a box-regression layer and a box-classification layer. Different from RPN~\cite{ren2015faster}, we incorporate the features learned from the semantic segmentation task which can provide better local cues for objectness prediction and proposal localization. Specifically, we concatenate the semantic segmentation feature maps produced by the semantic segmentation network and the last shared convolutional feature maps along the channel axis, forming the input of the proposal generation network. We minimize an objective function following the multi-task loss in~\cite{ren2015faster} to optimize the parameters of the network.
\subsubsection{Recursive Detection Network}
The structure of the recursive detection network is based on the VGG-16 model~\cite{simonyan2014very}, which aims to score the input object proposals and refine their bounding box locations following the Fast R-CNN detection pipeline~\cite{girshick2015fast}. Different from Fast R-CNN, segmentation-aware features are constructed to incorporate guidance from the pixel-wise segmentation information which can help better depict the boundaries of the objects to facilitate detection. Specifically, the recursive detection network first utilizes an ROI pooling layer to generate a fixed-length feature descriptor of size $7\times7\times512$ from both the semantic segmentation feature maps and the last shared convolutional feature maps for each proposal provided by the proposal generation network. Then, following the feature combination scheme adopted in~\cite{bell2015inside}, we concatenate each pooled feature descriptor along the channel axis and reduce the dimension with a $1\times1$ convolution to match the shape of $7\times7\times512$ required by the first fully-connected layer (fc6) of the pre-trained VGG-16 model. To match the original amplitudes, each pooled feature map is L2 normalized and re-scaled back up by a fixed scale of $1000$. The generated feature is then fed into two fully-connected layers (fc6 and fc7) to predict the confidences over $K + 1$ categories, including $K$ object classes and one background class, as well as the bounding-box regression offsets. The parameters of these predictors are optimized by minimizing soft-max loss and smooth L1 loss~\cite{girshick2015fast}.

\subsection{Group Recursive Learning: An Expectation-Maximization Perspective}
The group recursive learning works in an expectation-maximization like way, where the network parameter learning and group recursive refinement are alternatively performed. In particular, in the maximization step, the network is trained to minimize the loss function or equivalently maximize the likelihood of multiple object bounding box predictions. In the expectation step, the locations of the proposals are refined with induced group information. We now proceed to provide more details about the EM-like group recursive learning.

\subsubsection{The M-Step: Mini-Batch Gradient Descent}
Specifically, the initial object proposal is denoted as $l$ where $l=(l_x,l_y,l_w,l_h)$ specifies its pixel coordinates of the center $(l_x,l_y)$ and its width and height in pixels $(l_h,l_w)$. Each ground-truth bounding box $l^*$ is specified in the same way: $l^* = (l^*_x,l^*_y,l^*_w,l^*_h)$. The bounding box regression targets $r^*$ are computed as $r^*=f(l,l^*)$ following the transformation strategy $f(\cdot)$ adopted in~\cite{girshick2013rich}, in which $r^*$ specifies a scale-invariant translation and log-space height/width shift relative to an object proposal. In the $t$-th iteration, the network takes the refined bounding boxes $l_{t-1}$ produced in the $(t-1)$-th iteration as input, and predicts bounding-box regression offsets, $r_{t,k}=(r_{t,k}^x, r_{t,k}^y, r_{t,k}^w, r_{t,k}^h)$ for each of the $K$ object classes, indexed by $k$, and the category-level confidences $p_t=(p_{t,0},...,p_{t,k})$ for $K+1$ categories. Each training proposal is labeled with a ground-truth class $g$ and a ground-truth bounding-box regression target $r^{*}_{t}$. We use a multi-task loss $J$ on each object proposal to jointly train for classification and bounding-box regression:
\begin{equation}\label{eqn:loss}
	J_{t}=J_{cls}(p_{t},g)+\mathbf{1}[g\geq 1]J_{loc}(r_{t,g},r^{*}_{t}),
\end{equation}
where $J_{cls}$ and $J_{loc}$ are the losses for the classification and the bounding-box regression, respectively. In particular, $J_{cls}(p_{t},g) = -\log p_{t,g}$ is log loss for the ground truth class g and $J_{loc}$ is a smooth $L_1$ loss proposed in~\cite{girshick2015fast}. The Iverson bracket indicator function $\mathbf{1}[g\geq 1]$ equals 1 when $g\geq 1$ and 0 otherwise. For background proposals (\emph{i.e. }$g=0$), the $J_{loc}$ is ignored. After the training process, the loss $J$ in the $t$-th iteration will be minimized and the likelihood of the regressed proposals being near to the corresponding ground truth is maximized. 

\subsubsection{The E-Step: Group Confidence Pooling}
The regressed bounding box $l_t$ of the proposal can be computed as $f^{^{-1}}(l_{t-1},r_{t,g})$, where $f^{^{-1}}(\cdot)$ represents the inverse operation of $f(\cdot)$. The final bounding box coordinates are further refined by considering the locations of all the surrounding proposals at different parts of the same object through a group confidence pooling scheme. Specifically, for a specific refined proposal $l_{t,i}$, denote $D_t$ as the set of proposals of the same class that have an overlap with $l_{t,i}$ of more than 0.7 on IOU metric. The refined location of $l_{t,i}$ can be taken as the expected location of the group by regarding the confidence score $s_{t,j}$ of each proposal $l_{t,j}\in D_t$ as a weight: 
\begin{equation}
	{l}'_{t,i}=\frac{\sum _{j:l_{t,j}\in D_t}s_{t,j}\cdot l_{t,j}}{\sum _{j:l_{t,j}\in D_t}s_{t,j}}.
\end{equation}
With this group confidence pooling scheme, the proposals will be refined to a better location by taking the surrounding proposals into consideration. The better localized proposals will be given higher confidence scores. As a result, both loss terms in Eqn.~\eqref{eqn:loss} will be reduced.

Both the M-step and the E-step optimization can be realized within an end-to-end framework. Assume that the total number of refinement iterations is $T$. During the optimization, we unroll the detection network  by stacking $T$ detection networks with \emph{shared} parameters. The global loss is computed as
\begin{equation}
	J=\sum _{t\leq T}J_{t}+J_{pgn},
\end{equation}
where $J_{t}$ (ref. Eqn.~\eqref{eqn:loss}) represents the loss produced by the recursive detection network at the $t$-th iteration with refined proposals and $J_{pgn}$ denotes the loss output by the proposal generation network following the multi-task loss in~\cite{ren2015faster}. Thus the multi-stage network cascades with group recursive learning can be trained end-to-end jointly.

\subsection{Testing}
In testing, given an input image, the proposed framework first generates initial object proposals using the proposal generation network and then recursively passes them into the recursive detection network. At the $t$-th iteration, the recursive detection network predicts the category-level confidences $p_t$ and bounding-box regression offsets $r_{t}$ for each proposal. The category of the proposal is predicted as the class with the maximum score in $p_t$. For the proposals predicted as a specific object class, the locations of the proposals are updated by refining the previous location $l_{t-1}$ with the predicted bounding-box regression offsets $r_{t,g}$ and then performing the group confidence pooling scheme as previously mentioned. For the proposals predicted as the background class, the locations of the proposals are not updated. The final outputs for each proposal are the results in the last iteration $t=T$, including the predicted category-level confidences $p_T$ and the refined locations $l_{T}$.

\begin{table*} [htbp]\setlength{\tabcolsep}{1.8pt}
	\centering\scriptsize
	{\caption{Detection results on VOC 2007 test. \textbf{P}: incorporate semantic features for object proposal generation, \textbf{D}: incorporate semantic features for object classification and bounding box regression, \textbf{R}: perform group recursive learning.} \label{tab:VOC2007}	
		\begin{tabular}{ccccccccccccccccccccccc}
			\toprule
			\textbf{Method} & \textbf{P D R}   \vline & \textbf{mAP} \vline& \tiny{\textbf{aero}} & \tiny{\textbf{bike}} & \tiny{\textbf{bird}} & \tiny{\textbf{boat}} & \tiny{\textbf{bottle}} & \tiny{\textbf{bus}} & \tiny{\textbf{car}} & \tiny{\textbf{cat}} & \tiny{\textbf{chair}} & \tiny{\textbf{cow}} & \tiny{\textbf{table}} & \tiny{\textbf{dog}} & \tiny{\textbf{horse}} & \tiny{\textbf{mbike}} & \tiny{\textbf{person}} & \tiny{\textbf{plant}} & \tiny{\textbf{sheep}} & \tiny{\textbf{sofa}} & \tiny{\textbf{train}} & \tiny{\textbf{tv}} \\
			\midrule
			FRCN~\cite{girshick2015fast} &\quad\quad\quad\vline & 70.0  \vline &
			\tiny{77.0} & \tiny{78.1} & \tiny{69.3} & \tiny{59.4} & \tiny{38.3} & \tiny{81.6} & \tiny{78.6} & \tiny{86.7} & \tiny{42.8} & \tiny{78.8} & 
			\tiny{68.9} & \tiny{84.7} & \tiny{82.0} & \tiny{76.6} & \tiny{69.9} & \tiny{31.8} & \tiny{70.1} & \tiny{74.8} & \tiny{80.4} & \tiny{70.4} \\
			RPN~\cite{ren2015faster} &\quad\quad\quad\vline & 73.2 \vline & 
			\tiny{76.5} & \tiny{79.0} & \tiny{70.9} & \tiny{65.5} & \tiny{52.1} & \tiny{83.1} & \tiny{84.7} & \tiny{86.4} & \tiny{52.0} & \tiny{81.9} & 
			\tiny{65.7} & \tiny{84.8} & \tiny{84.6} & \tiny{77.5} & \tiny{76.7} & \tiny{38.8} & \tiny{73.6} & \tiny{73.9} & \tiny{83.0} & \tiny{72.6} \\
			ResNet-101~\cite{he2015deep} &\quad\quad\quad\vline & 76.4 \vline & 
			\tiny{79.8} & \tiny{80.7} & \tiny{76.2} & \tiny{68.3} & \tiny{55.9} & \tiny{85.1} & \tiny{85.3} & \tiny{89.8} & \tiny{56.7} & \tiny{87.8} & 
			\tiny{69.4} & \tiny{88.3} & \tiny{88.9} & \tiny{80.9} & \tiny{78.4} & \tiny{41.7} & \tiny{78.6} & \tiny{79.8} & \tiny{85.3} & \tiny{72.0} \\
			MR-CNN~\cite{gidaris2015object} &\quad\quad\quad\vline & 78.2 \vline & 
			\tiny{80.3} & \tiny{84.1} & \tiny{78.5} & \tiny{70.8} & \tiny{68.5} & \tiny{88.0} & \tiny{85.9} & \tiny{87.8} & \tiny{60.3} & \tiny{85.2} & 
			\tiny{73.7} & \tiny{87.2} & \tiny{86.5} & \tiny{85.0} & \tiny{76.4} & \tiny{48.5} & \tiny{76.3} & \tiny{75.5} & \tiny{85.0} & \tiny{81.0} \\
			\midrule
			Ours (Baseline)   &$\quad\quad\quad$ \vline & 76.0  \vline & 
			\tiny{78.6} & \tiny{80.1} & \tiny{77.7} & \tiny{67.0} & \tiny{63.2} & \tiny{86.1} & \tiny{87.9} & \tiny{89.0} & \tiny{58.7} & \tiny{82.4} & 
			\tiny{70.6} & \tiny{84.7} & \tiny{87.1} & \tiny{76.9} & \tiny{79.0} & \tiny{47.2} & \tiny{75.4} & \tiny{70.6} & \tiny{82.5} & \tiny{74.7} \\	
			Ours   &$\surd\quad\quad$ \vline & 76.4  \vline & 
			\tiny{79.4} & \tiny{79.9} & \tiny{76.5} & \tiny{69.3} & \tiny{62.8} & \tiny{86.8} & \tiny{87.5} & \tiny{88.5} & \tiny{58.2} & \tiny{83.3} & 
			\tiny{71.4} & \tiny{84.7} & \tiny{85.2} & \tiny{78.9} & \tiny{78.8} & \tiny{49.1} & \tiny{77.2} & \tiny{70.5} & \tiny{83.8} & \tiny{75.4} \\			
			Ours   &$\surd\surd\quad$ \vline & 77.6  \vline & 
			\tiny{78.7} & \tiny{85.7} & \tiny{76.8} & \tiny{71.8} & \tiny{64.7} & \tiny{85.7} & \tiny{87.5} & \tiny{87.7} & \tiny{60.2} & \tiny{85.2} & 
			\tiny{72.5} & \tiny{87.0} & \tiny{86.7} & \tiny{79.6} & \tiny{79.3} & \tiny{48.8} & \tiny{76.6} & \tiny{77.2} & \tiny{84.3} & \tiny{75.9} \\				
			Ours   &$\surd\surd\surd$ \vline & 78.6  \vline & 
			\tiny{80.0} & \tiny{81.0} & \tiny{77.4} & \tiny{72.1} & \tiny{64.3} & \tiny{88.2} & \tiny{88.1} & \tiny{88.4} & \tiny{64.4} & \tiny{85.4} & 
			\tiny{73.1} & \tiny{87.3} & \tiny{87.4} & \tiny{85.1} & \tiny{79.6} & \tiny{50.1} & \tiny{78.4} & \tiny{79.5} & \tiny{86.9} & \tiny{75.5} \\			
			\bottomrule
		\end{tabular}
	}	
\end{table*}    

\begin{table*} [htbp]\setlength{\tabcolsep}{1.8pt}
	\centering\scriptsize
	{\caption{Detection results on VOC 2012 test. \textbf{P}: incorporate semantic features for object proposal generation, \textbf{D}: incorporate semantic features for object classification and bounding box regression, \textbf{R}: perform group recursive learning.} \label{tab:VOC2012}	
		\begin{tabular}{ccccccccccccccccccccccc}
			\toprule
			\textbf{Method} & \textbf{P D R}   \vline & \textbf{mAP} \vline& \tiny{\textbf{aero}} & \tiny{\textbf{bike}} & \tiny{\textbf{bird}} & \tiny{\textbf{boat}} & \tiny{\textbf{bottle}} & \tiny{\textbf{bus}} & \tiny{\textbf{car}} & \tiny{\textbf{cat}} & \tiny{\textbf{chair}} & \tiny{\textbf{cow}} & \tiny{\textbf{table}} & \tiny{\textbf{dog}} & \tiny{\textbf{horse}} & \tiny{\textbf{mbike}} & \tiny{\textbf{person}} & \tiny{\textbf{plant}} & \tiny{\textbf{sheep}} & \tiny{\textbf{sofa}} & \tiny{\textbf{train}} & \tiny{\textbf{tv}} \\
			\midrule
			FRCN~\cite{girshick2015fast} &\quad\quad\quad\vline & 68.4  \vline &
			\tiny{82.3} & \tiny{78.4} & \tiny{70.8} & \tiny{52.3} & \tiny{38.7} & \tiny{77.8} & \tiny{71.6} & \tiny{89.3} & \tiny{44.2} & \tiny{73.0} &
			\tiny{55.0} & \tiny{87.5} & \tiny{80.5} & \tiny{80.8} & \tiny{72.0} & \tiny{35.1} & \tiny{68.3} & \tiny{65.7} & \tiny{80.4} & \tiny{64.2} \\
			RPN~\cite{ren2015faster} &\quad\quad\quad\vline & 70.4 \vline & 	
			\tiny{84.9} & \tiny{79.8} & \tiny{74.3} & \tiny{53.9} & \tiny{49.8} & \tiny{77.5} & \tiny{75.9} & \tiny{88.5} & \tiny{45.6} & \tiny{77.1} & 
			\tiny{55.3} & \tiny{86.9} & \tiny{81.7} & \tiny{80.9} & \tiny{79.6} & \tiny{40.1} & \tiny{72.6} & \tiny{60.9} & \tiny{81.2} & \tiny{61.5} \\			
			FRCN+YOLO~\cite{redmon2015you} &\quad\quad\quad\vline & 70.4 \vline& 	
			\tiny{83.0} & \tiny{78.5} & \tiny{73.7} & \tiny{55.8} & \tiny{43.1} & \tiny{78.3} & \tiny{73.0} & \tiny{89.2} & \tiny{49.1} & \tiny{74.3} & 
			\tiny{56.6} & \tiny{87.2} & \tiny{80.5} & \tiny{80.5} & \tiny{74.7} & \tiny{42.1} & \tiny{70.8} & \tiny{68.3} & \tiny{81.5} & \tiny{67.0} \\	
			HyperNet &\quad\quad\quad\vline & 71.4 \vline&                    
			\tiny{84.2} & \tiny{78.5} & \tiny{73.6} & \tiny{55.6} & \tiny{53.7} & \tiny{78.7} & \tiny{79.8} & \tiny{87.7} & \tiny{49.6} & \tiny{74.9} & 
			\tiny{52.1} & \tiny{86.0} & \tiny{81.7} & \tiny{83.3} & \tiny{81.8} & \tiny{48.6} & \tiny{73.5} & \tiny{59.4} & \tiny{79.9} & \tiny{65.7} \\
			ResNet-101~\cite{he2015deep} &\quad\quad\quad\vline & 73.8 \vline& 	
			\tiny{86.5} & \tiny{81.6} & \tiny{77.2} & \tiny{58.0} & \tiny{51.0} & \tiny{78.6} & \tiny{76.6} & \tiny{93.2} & \tiny{48.6} & \tiny{80.4} & 
			\tiny{59.0} & \tiny{92.1} & \tiny{85.3} & \tiny{84.8} & \tiny{80.7} & \tiny{48.1} & \tiny{77.3} & \tiny{66.5} & \tiny{84.7} & \tiny{65.6} \\		
			MR-CNN~\cite{gidaris2015object} &\quad\quad\quad\vline & 73.9 \vline& 	
			\tiny{85.5} & \tiny{82.9} & \tiny{76.6} & \tiny{57.8} & \tiny{62.7} & \tiny{79.4} & \tiny{77.2} & \tiny{86.6} & \tiny{55.0} & \tiny{79.1} & 
			\tiny{62.2} & \tiny{87.0} & \tiny{83.4} & \tiny{84.7} & \tiny{78.9} & \tiny{45.3} & \tiny{73.4} & \tiny{65.8} & \tiny{80.3} & \tiny{74.0} \\		
			\midrule
			Ours   &$\surd\surd\surd$ \vline & 74.9  \vline& 
			\tiny{85.7} & \tiny{82.0} & \tiny{75.0} & \tiny{62.7} & \tiny{58.3} & \tiny{80.5} & \tiny{80.3} & \tiny{89.4} & \tiny{55.8} & \tiny{78.2} & 
			\tiny{62.7} & \tiny{87.2} & \tiny{83.2} & \tiny{84.3} & \tiny{82.7} & \tiny{53.4} & \tiny{76.0} & \tiny{67.5} & \tiny{83.7} & \tiny{70.4} \\
			\bottomrule
		\end{tabular}
	}	
\end{table*} 

\section{Experiments}
\subsection{Experimental Settings}

\paragraph{Datasets and Evaluation Metrics}
To make fair comparison with the state-of-the-art methods~\cite{girshick2015fast}~\cite{ren2015faster}~\cite{gidaris2015object}, we evaluate the proposed framework on the PASCAL VOC 2007 benchmark and PASCAL VOC 2012 benchmark~\cite{everingham2010the}. The two datasets consist of 9,963 and 22,531 images respectively, and they are divided into train, val and test subsets. The model evaluated on VOC 2007 is trained based on the trainval split from VOC 2007, including 5,011 images, and the trainval split from VOC 2012, including 11,540 images. The model evaluated on VOC 2012 is trained based on all images from VOC 2007, including 9,963 images, and the trainval split from VOC 2012. We use standard evaluation metrics Average Precision (AP) and mean of AP (mAP) following the PASCAL challenge protocols for evaluation.

\paragraph{Implementation Details}
We initialize the bottom shared convolutional layers and the recursive detection network with the pre-trained VGG-16 model~\cite{simonyan2014very} and initialize the semantic segmentation network with the pre-trained semantic segmentation-aware CNN model in~\cite{gidaris2015object}. All the other newly added layers are initialized by drawing weights from a zero-mean Gaussian distribution with standard deviation 0.01 and 0.001. Our code is based on the publicly available Faster R-CNN framework~\cite{ren2015faster} built on the Caffe platform~\cite{jia2014caffe}. We fine-tune the whole framework jointly following the fine-tuning strategy proposed in~\cite{ren2015faster}. During fine-tuning, images are randomly selected for horizontally flipping with a probability of 0.5 to augment the training data. We set the iteration number for group recursive learning as $T = 2$, since only minor improvement with more iterations is observed. We run Stochastic Gradient Descent (SGD) for totally $140k$ iterations to train the network parameters for VOC 2007 and VOC 2012. The initial learning rate of all layers is set as 0.001 and decreased to one tenth of the current rate after $80k$ iterations. The model is trained on a NVIDIA GeForce Titan X GPU and Intel Core i7-4930K CPU @ 3.40 GHz. 

\begin{table*} [htbp]\setlength{\tabcolsep}{1.8pt}
	\centering\scriptsize
	{\caption{Comparison of performance with several architectural variants of our proposed framework on VOC 2007 test.} 
		\label{tab:Recursive}	
		\begin{tabular}{cccccccccccccccccccccc}
			\toprule
			\textbf{Method} \quad\quad\vline& \textbf{mAP} \vline& \tiny{\textbf{aero}} & \tiny{\textbf{bike}} & \tiny{\textbf{bird}} & \tiny{\textbf{boat}} & \tiny{\textbf{bottle}} & \tiny{\textbf{bus}} & \tiny{\textbf{car}} & \tiny{\textbf{cat}} & \tiny{\textbf{chair}} & \tiny{\textbf{cow}} & \tiny{\textbf{table}} & \tiny{\textbf{dog}} & \tiny{\textbf{horse}} & \tiny{\textbf{mbike}} & \tiny{\textbf{person}} & \tiny{\textbf{plant}} & \tiny{\textbf{sheep}} & \tiny{\textbf{sofa}} & \tiny{\textbf{train}} & \tiny{\textbf{tv}} \\
			\midrule
			Iter\_1 \quad\quad\quad\vline& 77.6  \vline & 
			\tiny{78.7} & \tiny{85.7} & \tiny{76.8} & \tiny{71.8} & \tiny{64.7} & \tiny{85.7} & \tiny{87.5} & \tiny{87.7} & \tiny{60.2} & \tiny{85.2} & 
			\tiny{72.5} & \tiny{87.0} & \tiny{86.7} & \tiny{79.6} & \tiny{79.3} & \tiny{48.8} & \tiny{76.6} & \tiny{77.2} & \tiny{84.3} & \tiny{75.9} \\			
			Iter\_2 \quad\quad\quad\vline & 78.6  \vline & 
			\tiny{80.0} & \tiny{81.0} & \tiny{77.4} & \tiny{72.1} & \tiny{64.3} & \tiny{88.2} & \tiny{88.1} & \tiny{88.4} & \tiny{64.4} & \tiny{85.4} & 
			\tiny{73.1} & \tiny{87.3} & \tiny{87.4} & \tiny{85.1} & \tiny{79.6} & \tiny{50.1} & \tiny{78.4} & \tiny{79.5} & \tiny{86.9} & \tiny{75.5} \\			
			Iter\_2\_testing \vline& 77.8 \vline&                    
			\tiny{80.0} & \tiny{80.8} & \tiny{76.7} & \tiny{72.4} & \tiny{63.9} & \tiny{85.4} & \tiny{88.0} & \tiny{89.1} & \tiny{59.5} & \tiny{85.1} & 
			\tiny{74.5} & \tiny{86.7} & \tiny{86.7} & \tiny{79.9} & \tiny{79.5} & \tiny{49.9} & \tiny{77.6} & \tiny{78.1} & \tiny{85.2} & \tiny{76.0} \\
			\bottomrule
		\end{tabular}
	}	
\end{table*}

\subsection{Performance Comparisons}
Table~\ref{tab:VOC2007} and Table~\ref{tab:VOC2012} provide the comparisons of the proposed framework with several state-of-the-art methods~\cite{girshick2015fast}~\cite{ren2015faster}~\cite{gidaris2015object}~\cite{redmon2015you}. It can be observed that our method obtains the mAP score of $78.6\%$ on VOC 2007, which outperforms the two baselines by $8.6\%$ for Girshick \emph{et al.}~\cite{girshick2013rich} and $5.4\%$ for Ren \emph{et al.}~\cite{ren2015faster}. On VOC 2012, our method outperforms the two baselines: $74.9\%$ vs $68.4\%$ of Girshick \emph{et al.}~\cite{girshick2013rich} and $70.4\%$ of Ren \emph{et al.}~\cite{ren2015faster}. In general, the proposed method shows significantly higher performance compared with the baselines and achieves competitive results compared with the state-of-the-art methods on both datasets, which validates its superiority in accurate object detection benefited from the multi-stage network cascades framework and the group recursive learning strategy.

\subsection{Ablation Studies}
We further evaluate two important components, \emph{i.e. }multi-stage network cascades and group recursive learning, to validate their effectiveness.

\paragraph{Multi-stage Network Cascades}
We verify the effectiveness of incorporating semantic segmentation features for better object proposal generation and detection using the multi-stage network cascades framework. As shown in Table~\ref{tab:VOC2007}, $0.4\%$ improvement can be observed by incorporating the semantic segmentation features into the proposal generation network compared to the variant without using semantic segmentation features where object proposals and detection results are directly generated based on the last shared convolutional features. Similarly, incorporating the semantic segmentation features into the object detection network offers a further performance increase of $1.2\%$. This demonstrates that the proposed multi-stage network cascades framework can effectively leverage the learned features from the semantic segmentation task for object detection, which leads to more accurate bounding boxes for object proposals and provides useful local cues for better object classification and localization.

\begin{figure*}
	\centering
	\includegraphics[scale=0.478]{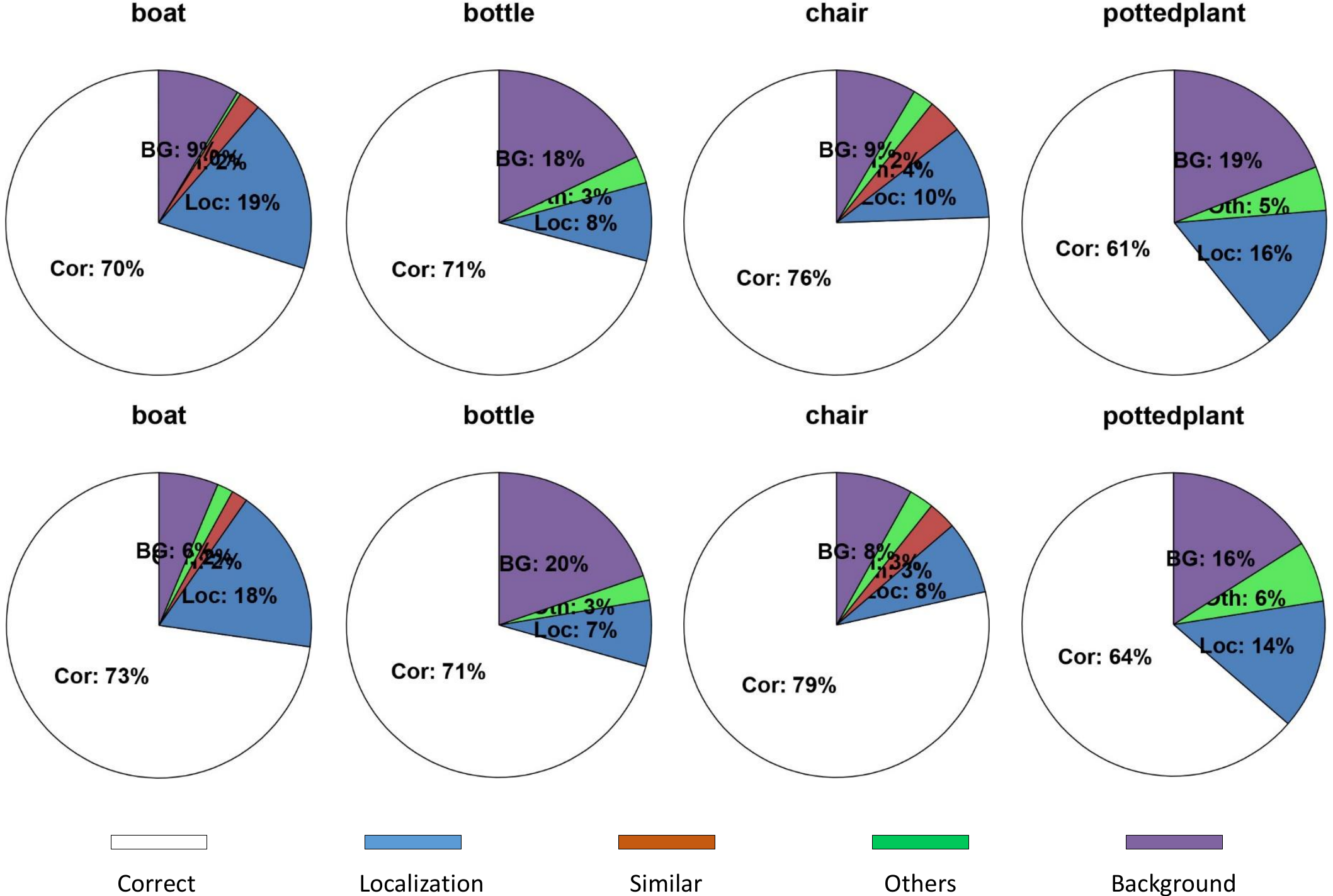}
	\caption{Analysis of top ranked false positives on VOC 2007 \emph{test}. Fractions of top $N$ detections ($N$ is the number of objects in the category) that are correct (Cor), or false positives due to poor localization (Loc), confusion with similar objects (Sim), confusion with other VOC objects (Oth), or confusion with background or unlabeled objects (BG), are shown.  We only show the graphs for challenging classes, \emph{i.e. }\emph{boat}, \emph{bottle}, \emph{chair} and \emph{pottedplant}, due to space limitations. \textbf{Top row:} the results of our baseline model. \textbf{Bottom row:} the results of the proposed method.}
	\label{fig:analysis1}
\end{figure*}

\begin{figure*}
	\centering
	\includegraphics[scale=0.478]{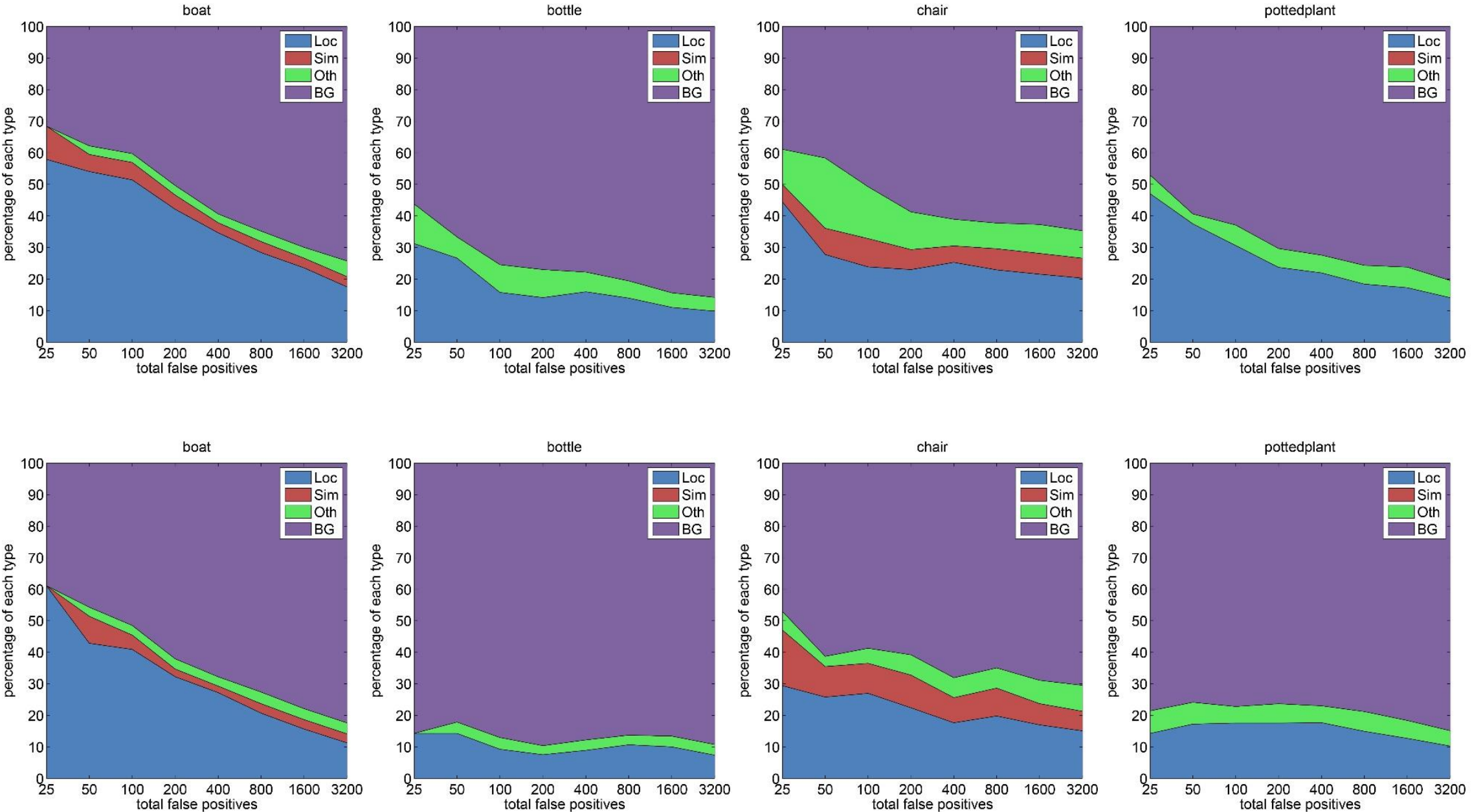}
	\caption{Top ranked false positive types on VOC 2007 \emph{test}. We only show the graphs for challenging classes, \emph{i.e. }\emph{boat}, \emph{bottle}, \emph{chair} and \emph{pottedplant}, due to space limitations. \textbf{Top row:} the results of the baseline model. \textbf{Bottom row:} the results of the proposed method.}
	\label{fig:analysis2}
\end{figure*}

\begin{figure*}
	\centering
	\includegraphics[scale=0.85]{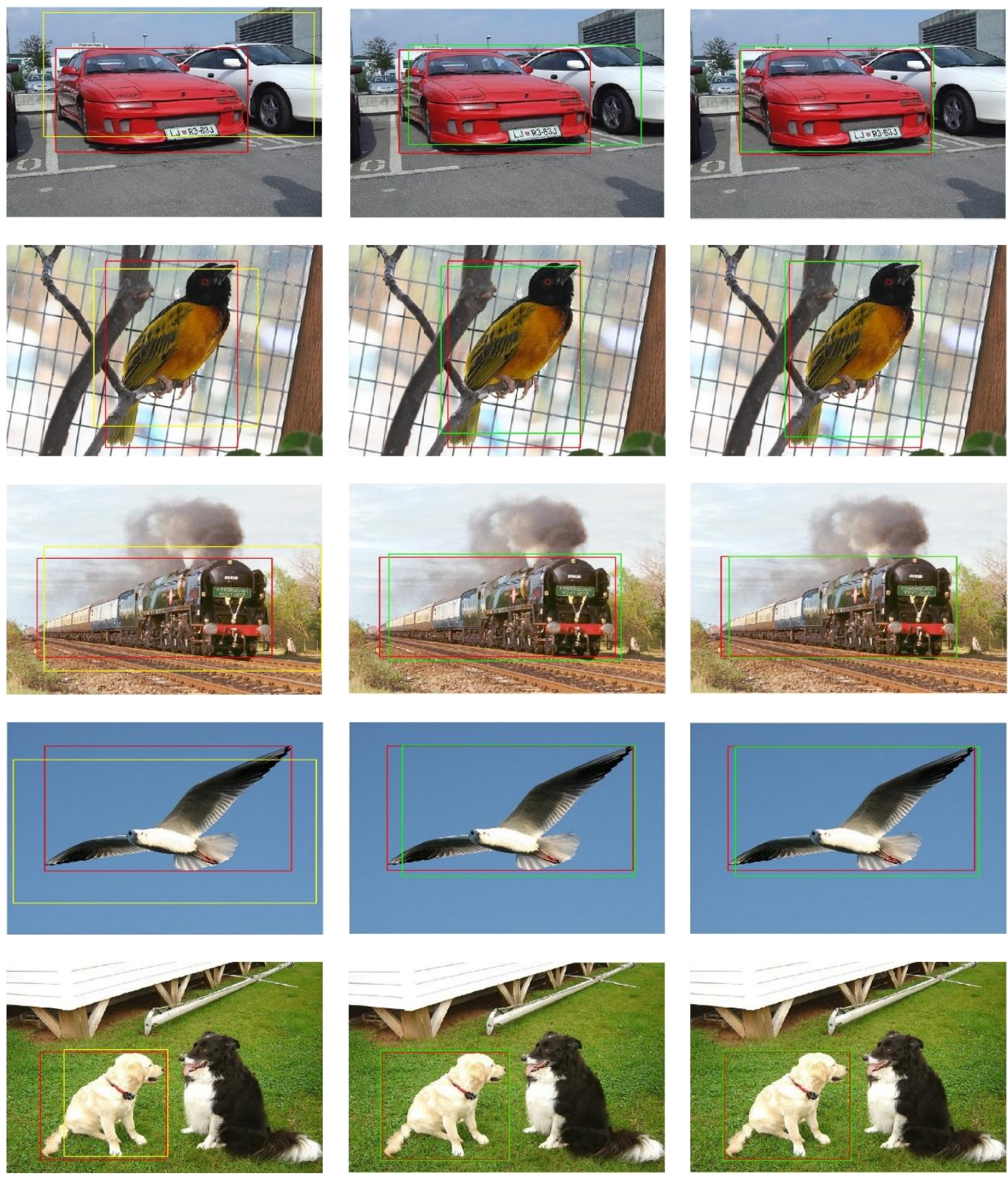}
	\caption{Qualitative results of the iterative bounding box location refinement procedure given an initial object proposal. The ground-truth bounding boxes of objects are annotated with red rectangles. The yellow and green rectangles represent the initial object proposal produced by the proposal generation network and the refined bounding box location from each refinement iteration, respectively.}
	\label{fig:visualization}
\end{figure*}

\paragraph{Group Recursive Learning}
In the proposed method, we set the maximal number of iterations for group recursive learning as $T=2$. To verify the effectiveness of the proposed group recursive learning scheme, we evaluate the performance of the proposed framework with different numbers of iterations during the training and testing stage. In Tabel~\ref{tab:Recursive}, ``Iter\_1" denotes the variant without using any recursive refinement where detection results are generated with only $1$ iteration and “Iter\_2" represents the model of using $2$ iterations. Compared with "Iter\_1", “Iter\_2” improves the performance by $1.0\%$, which verifies that more precise detection results can be obtained benefited from the recursively refined bounding box locations and classification scores. Since no noticeable improvement can be observed by adding more iterations, we use $2$ iterations for group recursive learning throughout our experiments.\\

To verify the advantage of using group recursive learning scheme in both the training and testing stage, we evaluate the performance of the variant where the recursive process is only performed during the testing stage, denoted as "Iter\_2\_testing". As shown in Tabel~\ref{tab:Recursive}, a $0.8\%$ drop in performance is observed by comparing "Iter\_2\_testing" with "Iter\_2",  demonstrating that employing group recursive refinement during both the training and testing stage is beneficial for jointly improving the network capabilities.

\subsection{Detection Error Analysis}
We analyze the detection errors of the proposed method using the tool of Hoiem \emph{et al.}~\cite{hoiem2012diagnosing}. In Figure~\ref{fig:analysis1}, we plot pie charts with the percentage of detections that are false positives due to bad localization, confusion with similar categories and other categories, and confusion with background or unlabeled objects. It can be observed that the proposed framework achieves a considerable reduction in the percentage of false positives due to bad localization for challenging categories. This validates that incorporating semantic segmentation features can increase the localization sensitivity of the detection network and precise bounding boxes for the detections can be obtained by adopting the proposed group recursive learning scheme. The similar observation can be deducted from  Figure~\ref{fig:analysis2} where we plot the top-ranked false positive types of the baseline and of the proposed framework.

\subsection{Qualitative Results}
In Figure~\ref{fig:visualization}, we provide sample qualitative results that present the iterative bounding box location refinement procedure starting from an initial object proposal produced by the proposal generation network. This example shows that our proposed method is capable of refining the produced initial object proposals step by step to fit them to the ground-truth bounding boxes of different objects, providing accurate object localization.

\section{Conclusion}
In this paper, we propose a multi-stage network cascades framework with group recursive learning for object detection. Specially, the proposed framework effectively utilizes semantic segmentation features to assist object detection by incorporating the semantic segmentation network, proposal generation network and recursive detection network into a unified architecture. In addition, a group recursive learning scheme is proposed to recursively score object proposals and regress their bounding boxes considering the locations of the surrounding proposals of the same object. We show that the proposed framework is particularly effective in object localization and achieves competitive results on PASCAL VOC 2007 and 2012.

\vspace{-0.05in}
\bibliographystyle{plain}
\scriptsize
\bibliography{tip}

\end{document}